\documentclass[conference]{IEEEtran}
\IEEEoverridecommandlockouts
\usepackage{cite}
\usepackage{amsmath,amssymb,amsfonts}
\usepackage{algorithmic}
\usepackage{graphicx}
\usepackage{textcomp}
\usepackage{xcolor}

\makeatletter

\def\ps@IEEEtitlepagestyle{
  \def\@oddfoot{\mycopyrightnotice}
  \def\@evenfoot{}
}
\def\mycopyrightnotice{
  {\footnotesize xxx-x-xxxx-xxxx-x/xx/\$31.00~\copyright~2018 IEEE\hfill} 
  \gdef\mycopyrightnotice{}
}

\@ifundefined{showcaptionsetup}{}{
 \PassOptionsToPackage{caption=false}{subfig}}
\usepackage{subfig}
\makeatother

\usepackage{eso-pic}
\newcommand\AtPageUpperMyright[1]{\AtPageUpperLeft{
 \put(\LenToUnit{0.5\paperwidth},\LenToUnit{-1cm}){
     \parbox{0.5\textwidth}{\raggedleft\fontsize{9}{11}\selectfont #1}}
 }}
\newcommand{\conf}[1]{
\AddToShipoutPictureBG*{
\AtPageUpperMyright{#1}
}
}

\newcommand{\fclass}{\mathcal{F}}
\newcommand{\vecX}{\mathbf{x}}
\newcommand{\vecT}{\boldsymbol{\theta}}

\def\BibTeX{{\rm B\kern-.05em{\sc i\kern-.025em b}\kern-.08em
    T\kern-.1667em\lower.7ex\hbox{E}\kern-.125emX}}
\begin{document}

\title{Effective training-time stacking for ensembling of deep neural networks}

\conf{2022 5th International Conference on
Artificial Intelligence and Pattern Recognition}

\author{\IEEEauthorblockN{1\textsuperscript{st} Polina Proskura}
\IEEEauthorblockA{\textit{Skoltech}\\
Russia \\
polina.231.11@gmail.com}
\and
\IEEEauthorblockN{2\textsuperscript{nd} Alexey Zaytsev}
\IEEEauthorblockA{
\textit{Skoltech}\\
Russia \\
a.zaytsev@skoltech.ru}

}

\maketitle

\begin{abstract}
Ensembling is a popular and effective method for improving machine learning (ML) models. It proves its value not only in classical ML but also for deep learning. Ensembles enhance the quality and trustworthiness of ML solutions, and allow uncertainty estimation. However, they come at a price: training ensembles of deep learning models eat a huge amount of computational resources. 

A snapshot ensembling collects models in the ensemble along a single training path.
As it runs training only one time, the computational time is similar to the training of one model. 

However, the quality of models along the training path is different: typically, later models are better if no overfitting occurs. So, the models are of varying utility.

Our method improves snapshot ensembling by selecting and weighting ensemble members along the training path.
It relies on training-time likelihoods without looking at validation sample errors that standard stacking methods do. Experimental evidence for Fashion MNIST, CIFAR-10, and CIFAR-100 datasets demonstrates the superior quality of the proposed weighted ensembles c.t. vanilla ensembling of deep learning models.
\end{abstract}


\section{Introduction}
The process of neural network training is cumbersome. 
The state-of-the-art models occupy significant volume and train for days or even weeks ~\cite{DBLP:journals/corr/LivniSS14,DBLP:journals/corr/abs-2102-00527}. 
The problem is even worse for ensembles of deep learning models trained in an independent way~\cite{breiman1996bagging}, as we need to run training several times.

More time-effective approaches to the training of deep learning model ensembles exist~\cite{snapshot,loss_surface}.
The snapshot ensembling methods collect ensemble members along the training path.
As we need only one training, we save resources with little drop in the quality of the final ensemble~\cite{chirkova2020deep}. 
One can show, that these separate models are diverse enough to reduce the bias of the estimator, as they come from different local optima of a loss function. 

The goal of the article is to propose an approach that works at the upper weighting level on a training time of single model and outperforms every single model. By looking at training time likelihood we were able to identify weights of separate ensemble members that lead to a better ensemble.

We evaluate our ensembling approaches for images classification problems~\cite{loss_surface} and investigate our ensembles with respect to the quality, diversity, and effectiveness.
We show, that our stacking procedure produces better models, than the vanilla snapshot ensembling, while making them more efficient.

Our work has the following sections:
\begin{itemize}
    \item We review the state-of-the-art in Section~\ref{sec:sota}.
    \item We propose our stacking procedure in Section~\ref{sec:methods}.
    \item We finish with the results of computational experiments in Section~\ref{sec:experiments}.
\end{itemize}
\section{State of the art}
\label{sec:sota}

\subsection{Ensembling in Machine learning}
For classical machine learning (ML) algorithms, ensembling is a very effective method for model improvement, getting rid of overfitting and decreasing uncertainty. 
Basic approaches average model predictions reducing ensemble model bias~\cite{randfor,rf3}, while gradient boosting approaches try to reduce both bias and variance~\cite{gradboost,gr2,adaboost,ab2,ab3} by adding basic models that correct errors of the current ensemble. For Random forest the other type of ensembling is used -- bagging~\cite{breiman1996bagging}.

\subsection{Stacking}
Stacking is one of the ways of ensembling \cite{stack_first}.  The main idea of the stacking is that the combiner algorithm is used on top of regular algorithms to make a final prediction using the predictions of the trained algorithms as a new input \cite{stack_two}.

The classical approach to the stacking in neural network ensembling is to train individually several neural networks and then use major voting principles for the calculation of the final result~\cite{stack_three}.

\subsection{Ensembles of neural networks}

A typical neural network has billions of parameters. 
So, if we optimize the loss function for a similar dataset and architecture with approaches based on stochastic gradient descent, we end up in different local optima due to the complexity of the loss function surface~\cite{DBLP:journals/corr/abs-2104-02395,SENDI2019271,58871}.
So, the total quality of the ensemble will be better than the quality of individual models~\cite{chirkova2020deep}. 
 
This solution increased the quality but requires more time to train an ensemble~\cite{DBLP:journals/corr/LivniSS14,DBLP:journals/corr/abs-2102-00527}. 
In \cite{powerlaw}, the authors explore dependencies of quality on parameters of the ensemble, stating that ensembles can provide better accuracy in both quality and time efficiency, than a single model after careful selection of architecture of the model.

\subsection{Effective ensembles of neural networks}

The authors \cite{cyclical} suggest to effectively ensemble models along the training path: cyclical learning rate \cite{8517060}. This method varies the learning rate during the training process~\cite{9232482}. 
The experiments confirmed the effectiveness of such methods~\cite{8517060,large}. 
Another paper on the application of this idea to ensembling of neural networks is \cite{snapshot}.

The training time is equal to the training time of one model, which means the significant economy of computer resources.
The authors of the articles suggest the schedule for the learning rate during training.

\subsection{Conclusions}
Ensembling methods help to achieve better results compared to a single model for both classic machine learning and deep learning ensembles.
For deep learning, efficient ensembling happens via snapshot ensembling: we collect models in the ensemble during a single training run with a cyclical learning rate.

However, the snapshot ensembling heuristic lacks a procedure to aggregate basic models into the ensemble, and the published results that these basic models vary in quality and diversity. 

\cite{snapshot,izmailov2018averaging} straightforwardly averages individual models into ensemble.
Bayes and stacking perspective at this type of ensembles suggest that there should exist a way to improve an ensemble.

\section{Methods}
\label{sec:methods}

\subsection{Background}

We consider a fixed architecture of a neural network. 
The architecture defines a set of functions $\fclass = \{f(\vecX)\}$ that comprises the space of the basic models for an ensemble. 
An example of the input $\vecX$ is an image with several channels, an example of the output $f(\vecX)$ is a probability vector with the size $k$ for the classification problem with $k$ classes from a set $\{1, \ldots, k\}$.
The vector of the neural network parameters $\vecT$ defines a function $f_{\vecT}(\vecX) \in \fclass$. 
We will use $\vecT_i$ to define a model $f_i = f_{\vecT_i}$. 

Typically, to produce a model one estimates parameters $\vecT$ via minimizing an empirical loss function for a sample $D = \{(\vecX_j, y_j)\}_{j = 1}^m$.
The goal is to build an ensemble of $N$ models $\{\vecT_1, \ldots, \vecT_N\}$, and the rule for aggregating the predictions of separate models for gaining the ensemble prediction $\hat{f}(\vecX)$.

\subsubsection{Ensembling of independent models.} The basic independent algorithm consists of two steps:
\begin{enumerate}
    \item We train each model from an ensemble using a sample $D$ to get $f_{\vecT_i}(\vecX)$.
As the training of each model starts from a different initialization of $\vecT$, the resulting models are also different.
    \item The prediction of the ensemble is the mean of all models 
$\hat{f}(\vecX) = \frac{1}{N} \sum_{k = 1}^N f_{\vecT_k}(\vecX)$.
\end{enumerate}

If the cost of training one model is $T$, then the cost of the straightforward training the ensemble is $N T$. 

\subsubsection{Snapshot ensembling.} 
More computationally effective approach is \emph{Snapshot} ensembling~\cite{snapshot}.
Suppose, that we have $T$ epochs for a training of a model.
Then the learning rate on $t$-th epoch will be $\alpha(t)$.
We select the learning rate scheduler: $\alpha(t)$ cyclically changes from $\alpha_{min}$ to $\alpha_{max} > \alpha_{min}$.

At the $\alpha_{min}$ we reach the local minimum of the loss function and further by increasing the learning rate we leave the current local minimum and reach the next one ~\cite{snapshot}.
As a $k$-th model we will choose $\vecT_k$, which is the estimation of the parameters on the $k$-th minimum of the learning rate, so that $\alpha(t) = \alpha_{min}$ at time $k$.
The ensemble prediction is the mean of the base models prediction:
$\hat{f}(\vecX) = \frac{1}{N} \sum_{k = 1}^N f_{\vecT_k}(\vecX)$.
The training cost, in this case, is $T$ and is similar to the training cost for training of a single model. 

\subsubsection{SWA approach} The main idea of SWA (Stochastic Weight Averaging)~\cite{izmailov2018averaging} is that the ensembling happens not in the model space, but in the model parameters space. The method uses the combination of the model weights at the different stages of the training. 
So, the following two properties are achieved:
\begin{itemize}
    \item If we combine weights, totally there is one model. It speeds up inference.
    \item The training process has the same estimated complexity but leads to finding a better local minimum with smoother slopes. 
\end{itemize}

\subsection{Proposed stacking approach}

We propose a procedure for the stacking based on effective Snapshot ensembling~\cite{snapshot}.
We provide a more precise selection of the iteration for finding the $\vecT_k$ in the neighborhood of the $k$-th minimum of learning rate.
We also argue, that as earlier models are worse than later ones, we should reduce weights for them. In particular, we consider weighting predictions in an ensemble, such that:
\[
    \hat{f}(\vecX) = \frac{1}{N} \sum_{k = 1}^N w_k f_{\vecT_k}(\vecX),
\]
the weights choosing procedure $w_k$ will follow.

\subsubsection{Model collection}

For a typical number of epochs $350$ with number of iterations $100$ per epoch we have more than $3000$ models during training.
We can't use them all due to computational restrictions and shouldn't, 
as neighbor models are similar to each other and add little value if we take both of them instead of just one.
Moreover, for cyclic learning rates models obtained at high learning rates typically are of worse quality, so if included in the ensemble, they will decrease the ensemble quality.

\subsubsection{Models in the minimum of the cyclical learning rate} \label{ssub:min}

The basic approach~\cite{snapshot} collects models at the minima of the cyclical learning rate, as they correspond to a neighborhood of a local optima of the loss function.

\subsubsection{Models in the average value of learning rate} \label{ssub:middle}

To increase the diversity of an ensemble, we can also add to the ensemble models picked up at points with $\alpha = \frac{\alpha_{max} + \alpha_{min}}{2}$. 
As during stacking, we weight each model, we can assign small weights for such model in case they decrease the overall ensemble accuracy.

\subsubsection{Choice of the models from the window} 
\label{ssub: window}

\begin{figure}[h]
    \centering
    \includegraphics[width=0.48\textwidth]{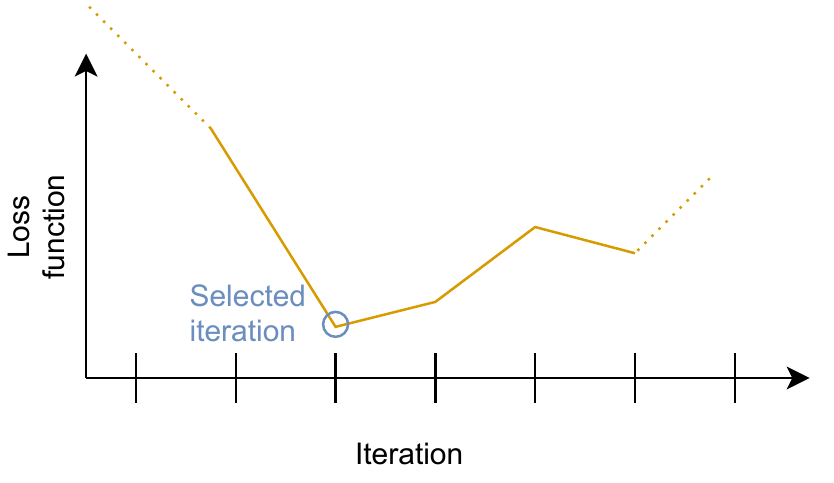}
    \caption{Choice of the model from the window near the minimum of learning rate}
    \label{fig:local_scheme}
\end{figure}

During the collection of models into the ensemble, we look at models at a local optimum of the learning rates.
However, we expect that this option can be not optimal.
To validate this assumption of snapshot ensemble, we propose the following change to the weight selection procedure illustrated by Figure ~\ref{fig:local_scheme}.
The steps are the following:
\begin{itemize}
    \item The local minimum of the learning rate occurs at $\alpha(t)$. We check, if models further on the path is better. A hyperparameter $s$ controls, that we take models in the range $(\alpha(t - s), \ldots, \alpha(t + s))$.
    \item For all $2 s + 1$ models around the local minimum we calculate the loss function. In particular, we use the loss function at the validation set and get $\{q_{t - s}, \ldots, q_{t + s}\}$.
    \item We select the model with the best $q_i$, $i \in \{{t - s}, \ldots, {t + s}\}$. We add this model to the ensemble.
\end{itemize}

\subsubsection{Weighting}

Snapshot ensembling has several disadvantages compared to classical ensembling.
One of them is the possible closeness of models in an ensemble, as we collect them during a single training. 
Contrary, in the classical ensembles due to randomization, different models end in different local minimums that are far from each other, so the different models make errors, which is not true for snapshot ensembles. 

As a solution to this challenge, we propose to use weighting. 
It helps to filter the models, which haven't reached the local minimum by giving them small weights, and to balance the effect of close models to each other in the ensemble. 

\subsubsection{Weighting based on the loss function value} \label{ssub:loss}

The main assumption of this approach is the following: the more errors the model makes, the less weight it should have.

Then the weighting is the following:
\begin{itemize}
    \item Let $l_i$ -- value of the loss function for the $i$-th model in ensemble.
    \item $w_i$ is the weight of the model can be calculated according the following formula:
    $$w_i = s(l_i)$$
    \item The function $s(\cdot)$ should reflect the following properties: 
    \begin{itemize}
        \item Be monotonous: the bigger the value of the loss function, the smaller the weight.
        \item Should be fast enough for calculations.
    \end{itemize}
\end{itemize}

We use the following simple function $s(l) = \frac{1}{l}$ to validate our approach.

\subsubsection{Weighting based on the likelihood on train and validation sets}
\label{ssub: likely}

The other automated approach has more freedom in the methodology -- finding the coefficients based on likelihood values for some samples.

Likelihood $L(\vecT)$ is the function which for the current parameters of the model $\vecT$, returns the probability that the sample $D$ comes from the distribution parametrized by $\vecT$: 
$$L(\vecT) = P(D | \vecT).$$

This approach can be described the following way:
\begin{itemize}
    \item Let $L_1, ... L_N$ be the likelihood for each from $N$ models in the ensemble calculated in the sample. 
    \item Then the weights of the model will be $w_1, ..., w_N$:
    $$w_i = s(L_i)$$
\end{itemize}

For the experiments was chosen the function $s(l) = l$.
The likelihood can be calculated on the training or validation set.

\subsubsection{Finding temperature for more optimal weighting of the models in ensemble} 
\label{ssub:temp}

This approach is the modification of the previous approach. The difference is that $s_{\tau}(\cdot)$ is another function, which has the temperature parameter $\tau$, responsible for weight calibration, to make the approach more flexible.
It has the following form
$s_{\tau}(l) = \exp{\left(\frac{l}{\tau}\right)}$.

Specific choices of temperature parameter $\tau$ can lead to a desired asymptotic behavior. 
With a big enough temperature, the weights will be almost the same and the prediction will be by simple averaging. 
With a small enough temperature the effect will be the opposite: the best model will have a big weight and others relatively small. 
The selection of a particular temperature allows a correct trade-off between these two extreme options.

\subsection{Implementation details}
\label{sec:implementation}

Most of the experiments were held using the big and complex neural network ResNet18 \cite{he2015deep}.
Neural network ResNet18 is based on convolutions with skip connections.

As the loss function, we use the negative logarithm of the likelihood: 
\[
Loss = - \log (P(D|\theta)), 
\]
where $D$ is a sample of the pairs of input and targets.

\section{Experiments}
\label{sec:experiments}

\subsection{Datasets}

We use three datasets for benchmarking: CIFAR-10~\cite{cifar10}, CIFAR-100~\cite{cifar10}, Fashion MNIST~\cite{DBLP:journals/corr/abs-1708-07747}.
All these datasets correspond to multiclass classification problems.

\subsection{Models in the minimum of the learning rate}

We start with checking the importance of temperature hyperparameters claimed to be important in previous articles.
We average models via likelihood values at the training sample.

\begin{figure}[h]
    \centering
    \includegraphics[width=0.5\textwidth]{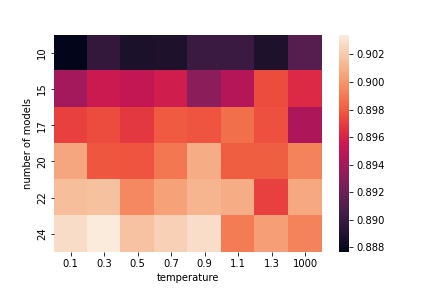}
    \caption{Results for CIFAR-10 for the averaging based on the likelihood on the training set.}
    \label{fig:heat_small}
\end{figure}

\begin{figure}[h]
    \centering
    \includegraphics[width=0.5\textwidth]{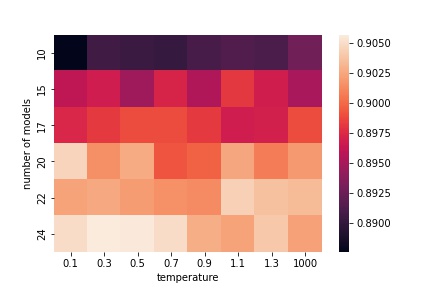}
    \caption{Results for CIFAR-10 for the averaging based on the likelihood on the validation set.}
    \label{fig:heat_new}
\end{figure}

Figure~\ref{fig:heat_small} provides results for different temperatures and a different number of models using weighting based on training likelihood values. 
Figure~\ref{fig:heat_new} uses the same procedure, but for the validation likelihood values used for the model weighting.
Value $1000$ on the temperature scale represents the results for the weighting with the equal weights. 

We see, that the best result is $90.5\%$ accuracy for a temperature near $1$ and a the maximum number of models, as expected for the training likelihood values and $90.4\%$ accuracy for ensemble based on validation likelihood values.
The results are similar for the weighting of ensembles using training, and test time errors are similar, so we can stack models to an ensemble using training data only.

\subsection{Models at the fixed distance from minimum}

In the following experiment we changed the iteration used to collect model: we use a base model not at the local optimum of the learning rate, but rather we wait fixed number of iterations.

\begin{figure}[h]
    \centering
    \includegraphics[scale=0.6]{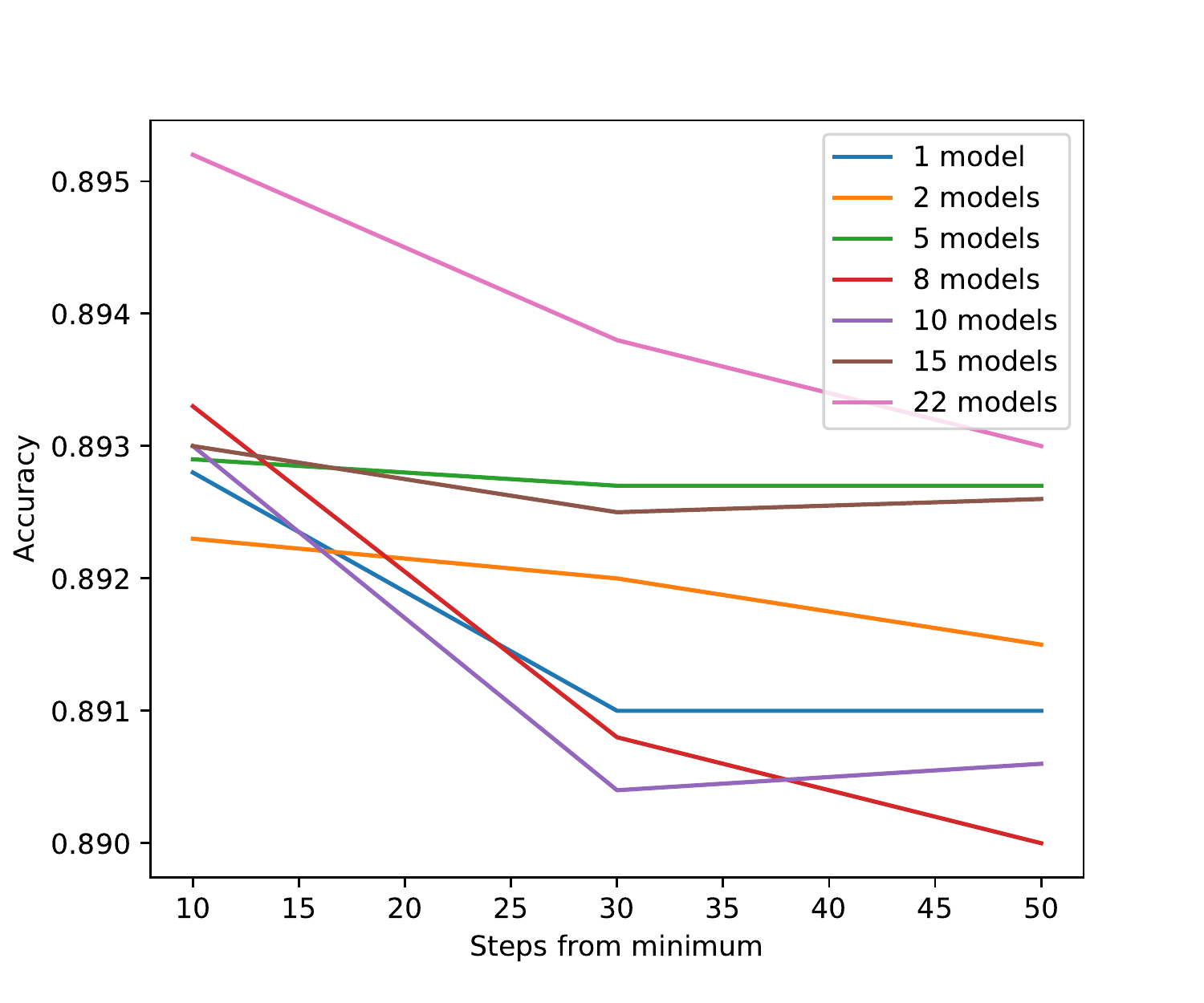}
    \caption{The dependency of the quality from the number of steps from the minimum learning rate, temperature = 1.0 for CIFAR-10}
    \label{fig:temp10}
\end{figure}

Results for the temperature $\tau = 1.0$ are in Figure~\ref{fig:temp10}. 

The best result is \textbf{89.5\%} occurs for $10$ steps difference from optimum.

\subsection{Main results}

We conclude our experiments by considering several options for snapshot ensembling model selection:
\begin{itemize}
    \item min -- models in the minimum learning rate, \emph{vanilla snapshot ensemble}.
    \item mid -- models in the minimum and the middle of the learning rate.
    \item $10$ steps -- model, which is in 10 steps from the minimum learning rate.
\end{itemize}

For model weighting, we consider two options:
\begin{itemize}
    \item eq -- ensemble with equal weights for all basic models.
    \item stack --- ensemble with weights obtained from training likelihood values
\end{itemize}

We also consider three baselines: a \emph{single} model performance, Stochastic weight averaging, and Ensemble that is an ensemble of independent models that takes prohibitive time for training, but can serve as a number to aim at for our effective ensembles based on snapshot ensembling.

Table~\ref{table:results} presents a comparison of the proposed options for snapshot ensembling and various baselines.
We see that the training time stacking in all cases improve the final results and is the most reasonable option to use. Moreover, using more models collected at the local maximum of the learning rate, we further improve the ensemble, as they provide more diversity to the model.

\begin{table}[h]
    \caption{Accuracy in percents for different combinations of the models with the best results for each ensembling type for CIFAR-10}
    \label{table:results}
    \centering
        \begin{tabular}{lllll}
        \hline
        Model & Type & Number & $\tau$ & Accuracy \\ 
        && of models && \\
        
        \hline
        Single & -- & 1 & -- & $89.49$ \\ 
        Ensembles & individual & 5 & -- & \textbf{91.9} \\
        \hline
         & min, eq & 24 & -- & $89.9$ \\
         & min, stack & 24 & 0.9 & $90.5$ \\
         & min, eq, val & 24 & -- & $90.25$ \\
        Snapshot & min, stack, val & 24 & 0.5 & $90.50$ \\
        ensemble & 10 steps, eq & 22 & -- & $89.0$ \\
         & 10 steps, stack & 22 & 1.0 & $89.55$ \\
         & middle, eq & 15 & -- & $89.9$ \\
         & middle, stack & 15 & 1.0 & \textbf{90.8} \\ 
        \hline
         Stochastic & min, eq & 24 & -- & $89.17$ \\
         Weight Averaging & min, stack & 24 & 1.0 & $89.23$ \\ 
        \hline
        \end{tabular}
\end{table}

\section{Acknowledgments}

Research was supported by Russian Science Foundation, grant 21-11-00373.

\section{Conclusion}
Cyclical learning rate helps to improve the quality of the model and more precisely detect the loss function minimums.
Snapshot ensembling with equal weights distribution gains the quality improvement compared to the single model but is worse than weighted ensembling and classical approach. 

The choice of the model for the snapshot ensembling in the minimum of the learning rate is logical and gains the most promising results.
Calculating the weights of the models in the ensemble according to the likelihood of every model on training and validation sample gains the best results among considered approaches and increases the quality of snapshot ensemble.

\bibliographystyle{ieeetr}
\bibliography{references}

\appendices
\section{Baseline results}

The table below~\ref{table:base_results} demonstrates the baseline results.
We see that Snapshot ensembles are better than individual models.
However, if we select a too large number of models, we end up with suboptimal ensembles.

 \begin{table}[h]
    \centering
        \begin{tabular}{lll}
        \hline
        Model & Amount of models & Accuracy, $\%$ \\ \hline
        Baseline  & 1  &$89.49$ \\ 
        Cyclical learning rate & 1 & $89.62$ \\
        \hline
        Ensemble & 3 & $91.3$ \\
         & 4 & $91.8$ \\
        & 5 & \textbf{$91.9$} \\
        \hline
        Snapshot ensemble & 3 & $89.95$ \\
        & 4 & $89.94$ \\ 
         & 5 & $89.88$ \\ \hline
        \end{tabular}
    \caption{Comparison of baseline models for CIFAR-10 data. Snapshot ensembles improve over baselines for a range of models in the ensemble}
    \label{table:base_results}
    \end{table}

\end{document}